\documentclass[10pt,twocolumn]{article}

\usepackage[letterpaper,margin=0.75in]{geometry}
\usepackage[T1]{fontenc}
\usepackage[utf8]{inputenc}
\usepackage[numbers,sort&compress]{natbib}
\usepackage{xcolor}
\usepackage{colortbl}
\usepackage{graphicx}
\usepackage{multirow}
\usepackage{amsmath,amssymb,mathtools,bm}
\usepackage{booktabs,tabularx,makecell}
\usepackage[hidelinks]{hyperref}
\usepackage{cleveref}

\AtBeginDocument{%
  }

\newcommand{\clip}{\operatorname{clip}}

\newcommand{\mean}{\operatorname{mean}}

\newcommand{\DWConv}{\operatorname{DWConv}}

\providecommand{\keywords}[1]{\par\noindent\textbf{Keywords: }#1}

\newcommand{\papertitle}{UHD-GPGNet: UHD Video Denoising via Gaussian-Process-Guided Local Spatio-Temporal Modeling}
\newcommand{\corrauthor}{Zhuoran Zheng}
\newcommand{\corremail}{zhengzr@njust.edu.cn}

\begin{document}

\twocolumn[
\begin{@twocolumnfalse}
\begin{center}
{\LARGE\bfseries \papertitle\par}
\vspace{0.9em}
{\large
Weiyuan He$^{1}$, Chen Wu$^{2}$, Pengwen Dai$^{3}$, Wei Wang$^{3}$, Dianjie Lu$^{4}$, Guijuan Zhang$^{4}$, Linwei Fan$^{5}$, Yongzhen Wang$^{6}$, Zhuoran Zheng$^{3}$\par}
\vspace{0.45em}
{\normalsize
$^{1}$Qingdao University, Qingdao, China\\
$^{2}$National University of Defense Technology, Changsha, China\\
$^{3}$SUN YAT-SEN University, Guangzhou, China\\
$^{4}$Shandong Normal University, Jinan, China\\
$^{5}$Shandong University of Finance and Economics, Yantai, China\\
$^{6}$Anhui University of Technology, Chuzhou, China\\
\par}
\vspace{0.45em}
{\normalsize\textbf{Corresponding author:} \corrauthor\ \texttt{<\corremail>}\par}
\vspace{0.9em}
\begin{minipage}{0.97\textwidth}
\begin{abstract}
Ultra-high-definition (UHD) video denoising requires simultaneously suppressing complex spatio-temporal degradations, preserving fine textures and chromatic stability, and maintaining efficient full-resolution 4K deployment. 
  In this paper, we propose UHD-GPGNet, a Gaussian-process-guided local spatio-temporal denoising framework that addresses these requirements jointly. 
  Rather than relying on implicit feature learning alone, the method estimates sparse GP posterior statistics over compact spatio-temporal descriptors to explicitly characterize local degradation response and uncertainty, which then guide adaptive temporal-detail fusion. 
  A structure-color collaborative reconstruction head decouples luminance, chroma, and high-frequency correction, while a heteroscedastic objective and overlap-tiled inference further stabilize optimization and enable memory-bounded 4K deployment. 
 Experiments on UVG and RealisVideo-4K show that UHD-GPGNet achieves competitive restoration fidelity with substantially fewer parameters than existing methods, enables real-time full-resolution 4K inference with significant speedup over the closest quality competitor, and maintains robust performance across a multi-level mixed-degradation schedule.A real-world study on phone-captured 4K video further confirms that the model, trained entirely on synthetic degradation, generalizes to unseen real sensor noise and improves downstream object detection under challenging conditions.
\end{abstract}
\vspace{0.4em}
\end{minipage}
\end{center}
\vspace{1.0em}
\end{@twocolumnfalse}
]

\section{Introduction}

\begin{figure}[t]
\centering
\includegraphics[width=\columnwidth]{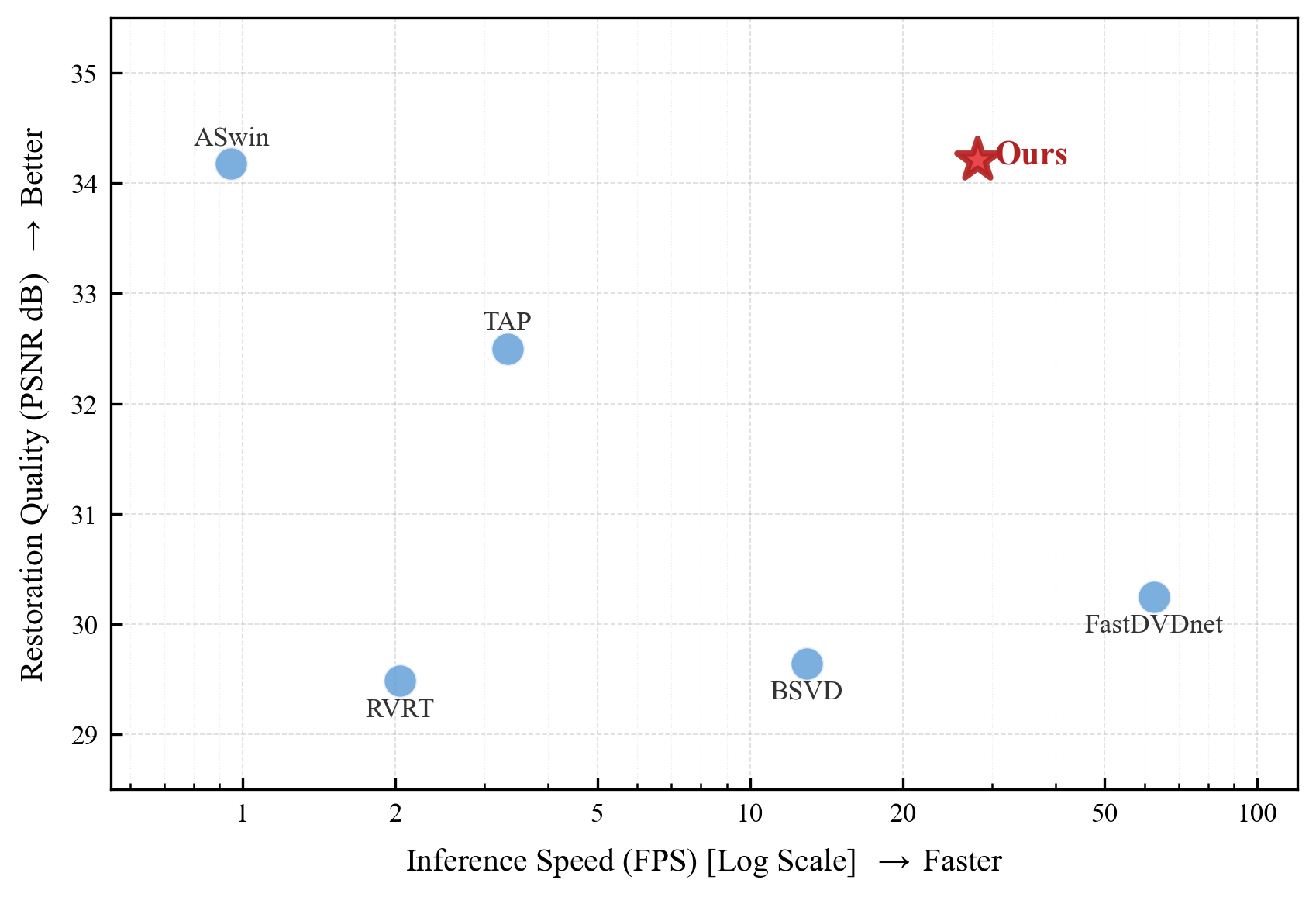}
\caption{Full-resolution 4K speed--quality--memory landscape. UHD-GPGNet~(\textcolor{red}{$\bullet$}) uniquely occupying the upper-right region. Smaller and higher-right is better.}
\vspace{-4mm}
\label{fig:speed_quality}
\end{figure}

Deep neural networks have driven rapid advances in video denoising~\cite{zhang2017beyond,chen2022simple,zamir2022restormer}, yet a substantial deployment gap persists when the target is an ultra-high-definition (UHD) video stream. 
A practical UHD denoiser must simultaneously suppress complex spatio-temporal degradations, preserve fine textures and chromatic stability under motion, and fit within the stringent memory and latency budget of $3840\times2160$ inference. 
High-speed inference is particularly challenging for video streams, where sensors may generate noise, and denoising models must perform full inference on continuous signals at full resolution~\cite{zheng2021ultra,li2023uhdfour,wang2024uhd}.

\begin{figure*}[t]
\centering
\includegraphics[width=\textwidth]{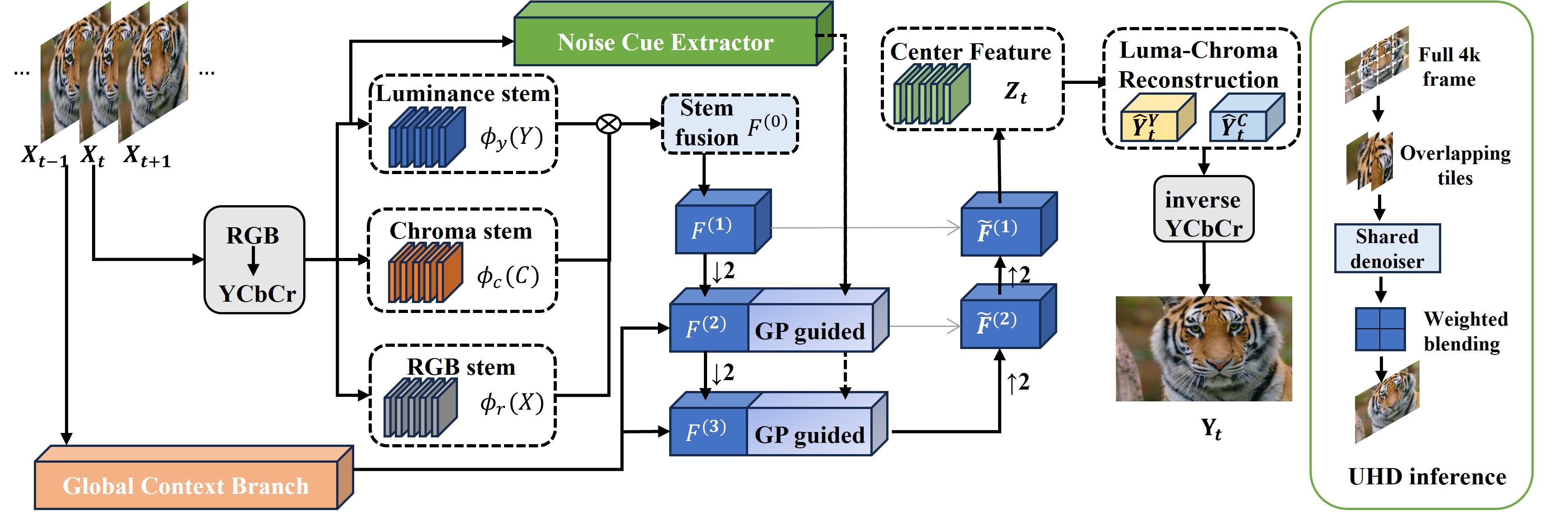}
\caption{Overall architecture of the proposed UHD video denoiser, including the Y/C/RGB stems, hierarchical multi-stage feature backbone, sparse GP-guided fusion blocks, structure-color reconstruction head, and full-resolution inference pipeline.}
\label{fig::overall_arch}
\vspace{-4mm}
\end{figure*}
Existing video denoisers struggle to strike a good trade-off between restored fidelity and deployment costs.
In terms of efficiency, compact designs such as FastDVDnet~\cite{tassano2020fastdvdnet}, BSVD~\cite{qi2022real}, and ASwin~\cite{lindner2023lightweight} demonstrate that flow-free temporal stacking, streaming-friendly buffering, and lightweight window attention can each deliver strong throughput, but at the expense of fidelity in fine structures and chromatic transitions. 
At the quality end, high-capacity restorers such as RVRT~\cite{liang2022recurrent} and VRT~\cite{liang2024vrt} combine recurrent propagation with transformer attention to improve restoration, yet their model scale and runtime make full-resolution 4K deployment difficult. Between these two extremes, the regime of high-fidelity, deployment-feasible UHD denoising remains largely unoccupied.

Existing video denoising methods typically rely on learned features to automatically determine how to combine information across frames. However, real UHD videos suffer from spatially varying noise levels: smooth regions like sky can be safely denoised with strong multi-frame averaging, while edges, textures, and fine details are easily damaged by over-smoothing and ghosting artifacts. Since these methods do not explicitly account for local noise characteristics, a one-size-fits-all temporal fusion strategy often under-filters some regions while over-smoothing others. To address this issue, we propose to explicitly estimate local noise behavior and uncertainty in a way that is efficient enough for high-quality 4K UHD video denoising.

Therefore, we propose \textbf{UHD-GPGNet}, a Gaussian-process-guided local spatio-temporal framework for practical UHD video denoising. 
Rather than treating Gaussian processes as a loose conceptual metaphor, our method performs explicit sparse GP posterior estimation on inducing-token neighborhoods within each encoder stage. The resulting local posterior mean and variance maps serve as lightweight stochastic summaries of degradation severity, which then modulate an uncertainty-conditioned gate that balances temporal aggregation against structure-preserving bypass. 
To further address the dominant UHD failure modes of detail loss, chroma drift, and tiling artifacts, the framework incorporates a luminance--chroma collaborative reconstruction head, a heteroscedastic restoration objective grounded in predictive uncertainty~\cite{kendall2017uncertainties,nix1994estimating}, and overlap-tiled full-resolution inference.

The main \textbf{contributions} are:
\textbf{1)} We propose a GP-guided local spatio-temporal denoising framework. It uses sparse inducing-token posterior estimation to explicitly capture local degradation response and uncertainty, which guide adaptive temporal-detail fusion, and works efficiently under a practical 4K compute budget.
\textbf{2)} We propose a UHD-oriented architecture that integrates luminance-chroma-aware extraction, uncertainty-conditioned fusion, and collaborative reconstruction of the structure-color. It achieves sharper luminance recovery and more stable chroma without heavy full-resolution processing.
\textbf{3)} Extensive experimental results demonstrate our method’s effectiveness and practicality. Tests include synthetic 4K dataset fidelity, robustness, full-resolution 4K deployment, mechanism validation, and real-world phone-captured 4K video tests with downstream detection, showing a favorable quality-efficiency balance in all scenarios.

\section{Related Work}

\noindent \textbf{Efficient and streaming video denoising.}
FastDVDnet~\cite{tassano2020fastdvdnet} established a sliding-window paradigm with a compact flow-free temporal stack. BSVD~\cite{qi2022real} extended this to online streaming via bidirectional buffers, and ASwin~\cite{lindner2023lightweight} replaced explicit stacking with lightweight shifted-window attention. TURTLE~\cite{ghasemabadi2024learning} showed that truncated causal history states improve the efficiency--quality balance, Classic Video Denoising~\cite{jin2025classic} revisited differentiable classical filters for robustness under noise mismatch, Maggioni~et~al.~\cite{maggioni2021efficient} demonstrated competitive multi-stage recurrent fusion, and PaCNet~\cite{vaksman2021patch} exploited non-local self-similarity via patch-matching. Despite their diversity, these methods rely entirely on implicit feature representations for fusion decisions, without explicit local uncertainty reasoning.This limits their ability to adapt fusion strength to spatially varying 
degradation severity, a critical requirement for UHD content where flat 
regions and fine textures coexist within the same frame

\noindent \textbf{High-capacity restoration method.}
RVRT~\cite{liang2022recurrent} and VRT~\cite{liang2024vrt} combine recurrent propagation with transformer attention for strong restoration at high computational cost.  Restormer~\cite{zamir2022restormer}, SwinIR~\cite{liang2021swinir}, Neighborhood Attention~\cite{hassani2023neighborhood}, and NAFNet~\cite{chen2022simple} explore efficient attention variants. For video tasks, BasicVSR~\cite{chan2021basicvsr}, BasicVSR++~\cite{chan2022basicvsr++}, EDVR~\cite{wang2019edvr}, and TAP~\cite{fu2024temporal} demonstrate strong propagation-based restoration. ConvIR~\cite{cui2024revitalizing}, MambaIRv2~\cite{guo2025mambairv2}, and SUPIR~\cite{yu2024scaling} further diversify the backbone landscape. However, the model complexity of these methods renders most of them impractical for full-resolution 4K deployment.

\noindent \textbf{For probabilistic restoration method.}
Raw video denoising work, including RViDeNet~\cite{yue2020supervised}, RViDeformer~\cite{yue2025rvideformer}, CBDNet~\cite{guo2019toward}, Unprocessing~\cite{brooks2019unprocessing}, Rethinking Noise Synthesis~\cite{zhang2021rethinking}, and Dancing Under the Stars~\cite{monakhova2022dancing} established the importance of physically grounded noise models and camera pipelines. The degradation pipelines in Real-ESRGAN~\cite{wang2021realesrgan} and BSRGAN~\cite{zhang2021designing} showed that coupling multiple corruption types markedly improves real-world robustness, directly informing our mixed-degradation design. Recent degradation-aware methods~\cite{zhang2024binarized,mao2025making} further broaden this perspective. From the probabilistic side, sparse GP approximations~\cite{titsias2009variational,snelson2005sparse,quinonero2005unifying}, deep kernel learning~\cite{wilson2016deep}, heteroscedastic uncertainty estimation~\cite{kendall2017uncertainties,nix1994estimating}, and deep ensembles~\cite{lakshminarayanan2017simple} provide practical routes to uncertainty-aware reasoning. Shift-Net~\cite{li2023simple} and UniDVD~\cite{li2022unidirectional} complement these efforts with efficient temporal modeling. Our method bridges these threads by employing sparse GP-guided local posterior summaries as lightweight uncertainty cues within a deployment-oriented UHD denoising architecture.

\section{Method}

\subsection{Problem Setup}
Given a noisy RGB video clip $X=\{X_{t-r},\ldots,X_{t+r}\}$ with $X_i \in [0,1]^{3\times H \times W}$, the goal is to recover the clean center frame $Y_t$ at UHD resolution ($H\!\times\!W = 2160\!\times\!3840$) via a network $\hat{Y}_t = f_\theta(X)$. Throughout, $[\cdot\,;\,\cdot]$ denotes channel-wise concatenation and $\odot$ denotes element-wise multiplication.
Two constraints distinguish the UHD setting: (i)~the full-resolution feature volume exceeds single-pass GPU memory, necessitating tiled inference; (ii)~UHD content exhibits spatially heterogeneous degradation where flat regions benefit from aggressive temporal fusion while edges and textures are sensitive to over-smoothing. To address both, our model predicts local uncertainty statistics for adaptive fusion control and heteroscedastic loss reweighting, interpreted as corruption-severity proxies rather than calibrated Bayesian posteriors.

\subsection{Architecture Overview}
The overall architecture, illustrated in Fig.~\ref{fig::overall_arch}, follows a hierarchical multi-stage encoder--decoder design augmented with sparse GP-guided fusion at deeper scales. The network processes five-frame clips through a three-stage feature pipeline (base width 32, stage depths $2$, $3$, $4$), where each stage consists of pseudo-3D blocks that factorize spatial and temporal convolutions for efficiency. GP-guided fusion blocks are inserted at the second and third stages, followed by two lightweight refinement stages for progressive feature recovery. To decouple structure-dominant restoration from chroma stabilization, RGB inputs are first converted into the YCbCr space. Let $Y$ and $C$ denote the luminance and chroma components of the input clip, respectively. The stem features are defined as
\begin{equation}
F^{(0)}
=
W_f
\left[
\phi_y(Y)\,;\,
\phi_c(C)\,;\,
\phi_r(X)
\right],
\label{eq:stem}
\end{equation}
where $\phi_y$, $\phi_c$, and $\phi_r$ denote the luminance, chroma, and RGB stem extractors, respectively, and $W_f$ is a $1\times1\times1$ fusion projection. In parallel, a lightweight global branch summarizes clip-level context $g \in \mathbb{R}^{C_g}$, and a noise-cue extractor computes local spatial--temporal variation maps from the luminance clip.

\subsection{Noise-Cue Extraction}
The cue extractor provides compact local degradation evidence from the luminance clip by computing horizontal, vertical, and temporal variation alongside local variance. Luminance is prioritized because most visible UHD structural errors are Y-channel-dominated; chroma stabilization is handled by the reconstruction head. After a shallow projection, the cue volume $C_{\mathrm{cue}}$ captures both local structure and instability:
\begin{equation}
C_{\mathrm{cue}}
=
\psi
\left(
\left[
\lvert \partial_x Y \rvert \,;\,
\lvert \partial_y Y \rvert \,;\,
\lvert \partial_t Y \rvert \,;\,
\operatorname{Var}_{3\times3}(Y)
\right]
\right).
\label{eq:cue}
\end{equation}

\subsection{Sparse Gaussian-Process-Guided Local Modeling}
\begin{figure*}[t]
\centering
\includegraphics[width=\textwidth]{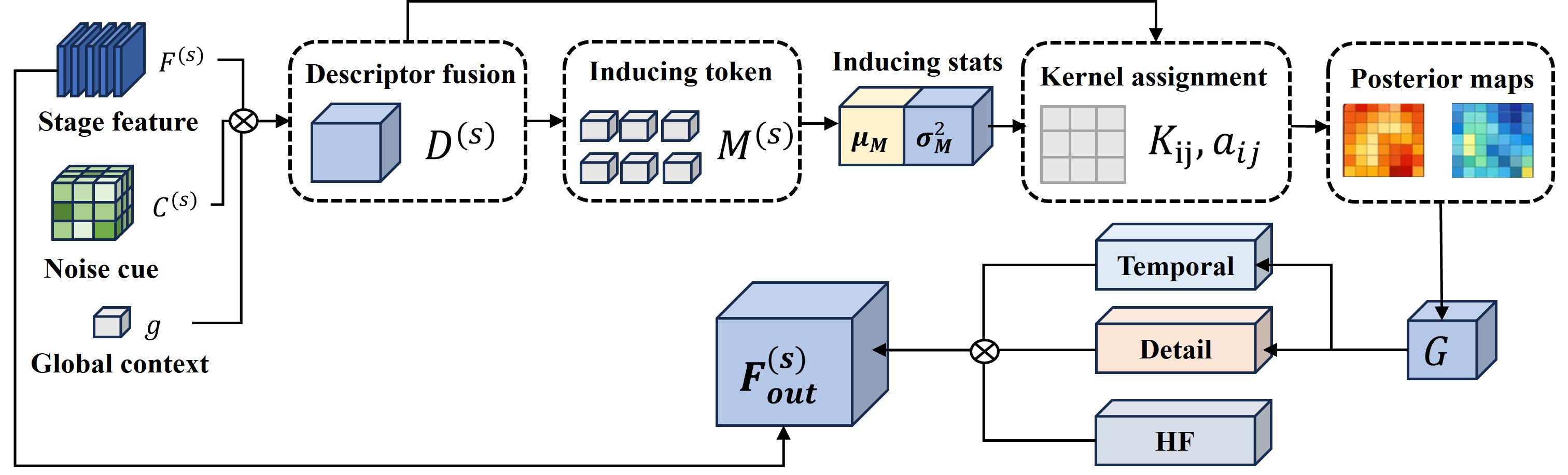}
\caption{Sparse GP-guided fusion module. The module forms local descriptors, pools inducing tokens, predicts inducing-level mean and variance statistics, produces local posterior mean and uncertainty maps, and converts them into fusion gates that balance temporal aggregation against detail-preserving bypass features.}
\label{fig:gp_module}
\end{figure*}

At selected encoder stages (Fig.~\ref{fig:gp_module}), we form a descriptor volume by combining the current feature map, the projected cue map, and the broadcast global context. For stage $s$, let $F^{(s)}$ denote the feature tensor and $C_{\mathrm{cue}}^{(s)}$ the resized cue tensor. The descriptor is defined as
\begin{equation}
D^{(s)}
=
W_d F^{(s)} + W_c C_{\mathrm{cue}}^{(s)} + W_g g .
\label{eq:descriptor}
\end{equation}
Let $d_i$ denote the flattened spatio-temporal tokens of the descriptor volume $D^{(s)}$. A lightweight pooling operator $P$ aggregates these dense descriptors into a compact inducing set, while a linear projection $W_q$ maps the dense descriptors to query tokens:
\begin{equation}
M^{(s)} = P(D^{(s)}) = \{m_j\}_{j=1}^{M},
\qquad
q_i = W_q d_i .
\label{eq:inducing}
\end{equation}
This separates a compact inducing set from the dense query lattice for stage-local probabilistic interaction. Each inducing token predicts a local mean $\mu_M$ and variance $\sigma_M^2$.

\emph{Design rationale.}
Unlike classical sparse GP methods such as FITC and VFE~\cite{snelson2005sparse,titsias2009variational}, we do not perform prior kernel matrix inversion or optimize inducing-point locations via marginal likelihood. Instead, inducing tokens are obtained by average-pooling on a fixed $4{\times}4$ spatial grid per frame, and all kernel parameters are optimized end-to-end via backpropagation. Our module borrows two key structural properties from sparse GP inference---kernel-weighted local aggregation and posterior moment decomposition via the law of total variance---and instantiates them as a differentiable, stage-local fusion primitive. This GP-structured design provides three specific \emph{inductive biases} absent in standard attention: (i)~the RBF length scale $\ell$ explicitly constrains the effective spatial aggregation radius, whereas dot-product attention determines receptive fields entirely implicitly; (ii)~the factored temporal parameter $\gamma$ allows temporal smoothing strength to be controlled independently of spatial similarity; (iii)~the variance in Eq.~\eqref{eq:posterior} is a mathematically exact total-variance decomposition under the kernel-weighted mixture, so high uncertainty directly reflects inducing-statistic disagreement rather than an arbitrary second-order feature. The controlled ablation in Table~\ref{tab:ablation_components} isolates this structural contribution from parameter-count effects.

Following the sparse inducing-point formulation~\cite{wilson2016deep,titsias2009variational,williams2006gaussian}, we define a stage-local spatio-temporal kernel between query $q_i$ and inducing token $m_j$:
\begin{equation}
K_{ij}
=
\alpha
\exp\!\left(
-\frac{\lVert q_i - m_j \rVert_2^2}{2\ell^2}
-\frac{(\tau_i-\tau_j)^2}{2\gamma^2}
\right),
\label{eq:kernel}
\end{equation}
where $\alpha$, $\ell$, and $\gamma$ are learned kernel parameters initialized to standard normal values and optimized jointly with all network weights, and $\tau_i$, $\tau_j$ encode normalized temporal positions. The kernel-based assignment weights are
\begin{equation}
a_{ij} = \operatorname{softmax}_j(K_{ij}).
\label{eq:assign}
\end{equation}
Using the inducing statistics, the local posterior moments become
\begin{equation}
\mu_i = \sum_j a_{ij}\mu_{M,j},
\qquad
\sigma_i^2
=
\sum_j a_{ij}\left(\sigma_{M,j}^2+\mu_{M,j}^2\right)-\mu_i^2 .
\label{eq:posterior}
\end{equation}
This follows the law of total variance: the posterior variance decomposes into within-component and between-component terms without matrix inversion; high uncertainty arises where inducing statistics disagree. For numerical stability, we form an uncertainty descriptor from the posterior response and log-variance:
\begin{equation}
u_i
=
\left[
\mu_i \,;\,
\log\!\left(\sigma_i^2+\varepsilon\right)
\right],
\qquad
\varepsilon > 0,
\label{eq:uncertainty_descriptor}
\end{equation}
where $\varepsilon$ prevents degenerate variance values.

\subsection{Uncertainty-Aware Temporal--Detail Fusion}
The core denoising decision is how much each location should trust temporally aggregated evidence versus a structure-preserving bypass path. We decompose the fusion into a temporal branch $T$, a detail-preserving spatial bypass $D$, and a high-frequency correction branch $H$:
\begin{align}
T(F)&=W_t^{\mathrm{pw}}\!\left(\DWConv_{3\times1\times1}(F)\right),\label{eq:Tbranch}\\
D(F)&=W_d^{\mathrm{pw}}\!\left(\DWConv_{1\times3\times3}(F)\right),\label{eq:Dbranch}\\
H(F)&=W_h^{\mathrm{pw}}\!\left(\delta\!\left(\DWConv_{1\times3\times3}(F)\right)\right).\label{eq:Hbranch}
\end{align}
The uncertainty-conditioned gate is defined as
\begin{equation}
G_i
=
\sigma\!\left(
W_2\,\delta\!\left(W_1 u_i\right)
\right).
\label{eq:gate}
\end{equation}
The fused output is
\begin{equation}
F_{\mathrm{out}}
=
F
+
W_o
\left[
G \odot T(F)\,;\,
(1-G)\odot D(F)
\right]
+
\beta\, H(F).
\label{eq:fusion}
\end{equation}
The gate is uncertainty-conditioned but not monotonically constrained, allowing context-dependent fusion: GP-guided statistics help distinguish regions benefiting from temporal aggregation from those requiring structure preservation.

\subsection{Structure--Color Collaborative Reconstruction}
The reconstruction head separates luminance-dominant restoration from chroma stabilization. The luminance and chroma outputs are predicted as
\begin{align}
\hat{Y}_t^{Y} &= \clip\!\left(X_t^{Y} + h_y(Z_t) + \alpha_h\, h_{\mathrm{hf}}(Z_t)\right),\label{eq:luma_head}\\
\hat{Y}_t^{C} &= \clip\!\left(X_t^{C} + \alpha_c\, h_c(Z_t)\right),\label{eq:chroma_head}
\end{align}
where $X_t^{Y}$, $X_t^{C}$ are the luminance and chroma components of the noisy center frame, $h_y$, $h_{\mathrm{hf}}$, $h_c$ are lightweight heads, and $\alpha_h$, $\alpha_c$ are learnable scales. The final RGB output is obtained by inverse YCbCr transform. This decomposition allows aggressive temporal smoothing on luminance while protecting chroma transitions and high-frequency detail.

\subsection{Noise Synthesis and Training Objective}
Following the high-order degradation paradigm~\cite{wang2021realesrgan,zhang2021designing}, we train with online mixed-degradation synthesis coupling exposure variation, signal-dependent sensor noise, read/grain noise, temporal flicker, chroma drift, blur, and optional H.264 compression:
\begin{equation}
\tilde{X}_t
=
C_{\mathrm{h264}}\!\Big(
T_{\mathrm{chr}}\!\big(
B(a_t X_t + n_{\mathrm{sen}}(X_t) + n_{\mathrm{read}} + n_g + \delta_t^{\mathrm{flicker}} + \delta_t^{\mathrm{chroma}})
\big)\Big).
\label{eq:noise_synth}
\end{equation}
Here $a_t$ is exposure gain, $n_{\mathrm{sen}}$/$n_{\mathrm{read}}$/$n_g$ are signal-dependent/read/grain noise, $B$ is blur, $C_{\mathrm{h264}}$ is H.264 compression, and $T_{\mathrm{chr}}$ is chroma perturbation.

The network predicts a log-variance map $s_t$ and is optimized with a heteroscedastic Charbonnier objective~\cite{kendall2017uncertainties,nix1994estimating} plus auxiliary regularization:
\begin{align}
L_{\mathrm{main}}
&=
\mean\!\left(
\rho(\hat{Y}_t,Y_t)\cdot \exp(-s_t)
+
\lambda_u s_t
\right),\label{eq:lmain}\\
L
&=
L_{\mathrm{main}}
+
\lambda_{\mathrm{hf}} L_{\mathrm{hf}}
+
\lambda_g L_{\mathrm{grad}}
+
\lambda_c L_{\mathrm{chr}}.\label{eq:loss}
\end{align}
The heteroscedastic term links uncertainty to optimization, while the auxiliary losses stabilize the dominant UHD failure modes of detail loss, edge softening, and chroma drift.

\subsection{Complexity and Tiled Inference}
The stage-level GP module reduces dense correlation cost from $\mathcal{O}(N^2)$ to $\mathcal{O}(NM)$ with $M \ll N$. Full-resolution 4K inference uses overlap-tiled restoration with normalized spatial windows:
\begin{equation}
\hat{Y}
=
{\sum_p w_p \odot f_\theta(X_p)}\big/{\sum_p w_p},
\label{eq:tiled}
\end{equation}
using $640{\times}640$ tiles with $64$-pixel overlap, avoiding seam artifacts while bounding peak memory.

\section{Experiments}
\subsection{Experimental Setup}

\noindent \textbf{Datasets.}
UVG~\cite{mercat2020uvg} serves as the main full-reference benchmark: original 4K frames are treated as pseudo-clean references and degraded online by the matched mixed protocol (Section~3.7). RealisVideo-4K~\cite{zhao2025realisvsrdetailenhanceddiffusionrealworld}, a detail-rich 4K source, is used for controlled cross-scene robustness testing under the same matched protocol.

\noindent \textbf{Baselines.}
We compare against FastDVDnet~\cite{tassano2020fastdvdnet}, ASwin~\cite{lindner2023lightweight}, BSVD~\cite{qi2022real}, RVRT~\cite{liang2022recurrent}, and TAP$^\dagger$~\cite{fu2024temporal}. All supervised baselines are initialized from official pretrained weights and fine-tuned on our mixed-degradation pipeline with matched training settings until convergence. Absolute PSNR of RVRT and TAP is lower than in their original papers due to the different noise distribution; the comparison is most informative for relative ranking under a common protocol. TAP$^\dagger$ is unsupervised and reported separately.

\noindent \textbf{Implementation details.}
All experiments use five-frame clips and center-frame restoration with a base width of 32, encoder depths $(2,3,4)$, and GP-guided blocks at stages 2--3 ($M{=}16$ inducing tokens per stage). Training runs for 80 epochs with AdamW (lr $2{\times}10^{-4}$, weight decay $10^{-4}$), mixed precision, gradient clipping at norm 1.0, batch size 4, and patch size $256{\times}256$. A 5-epoch warm-up uses milder degradation and reduced loss weights. Loss weights are $\lambda_u{=}0.02$, $\lambda_{\mathrm{hf}}{=}0.10$, $\lambda_g{=}0.05$, $\lambda_c{=}0.03$. Tiled inference uses $640{\times}640$ tiles with 64-pixel overlap. Deployment is profiled on an NVIDIA RTX PRO 6000 with synchronized CUDA timing (3 warm-up + 10 timed runs).

\begin{table}[t]
\caption{Main comparison on UVG. FLOPs are the estimated 4K equivalent ($640\!\times\!640$ per-tile cost $\times$ tile count); all methods use matched tiled inference.}
\label{tab:uvg_main}
\centering
\small
\setlength{\tabcolsep}{4pt}
\begin{tabular}{lcccc}
\toprule
Method & PSNR$\uparrow$ & SSIM$\uparrow$ & Params(M)$\downarrow$ & FLOPs(G)$\downarrow$ \\
\midrule
FastDVDnet         & 29.434 & 0.7675 & 2.479  & 334.4  \\
ASwin              & 32.764 & 0.9688 & 8.910  & 841.0  \\
BSVD               & 28.791 & 0.7178 & 9.816  & 3107.0 \\
RVRT               & 28.751 & 0.7316 & 12.787 & 888.9  \\
\midrule
TAP$^\dagger$      & 31.061 & 0.8977 & 34.482 & 3183.2 \\
\midrule
\textbf{UHD-GPGNet} & \textbf{32.759} & \textbf{0.9592} & \textbf{0.707} & \textbf{347.3} \\
\bottomrule
\end{tabular}
\end{table}

\subsection{Main Quantitative Comparison on UVG}
Table~\ref{tab:uvg_main} reveals a clear quality--complexity separation among the compared methods. UHD-GPGNet attains 32.759\,dB PSNR, essentially matching ASwin (32.764\,dB), while using only 0.707M parameters---$12.6\times$ fewer than ASwin and $18.1\times$ fewer than RVRT. This result demonstrates that near-frontier restoration fidelity is achievable at dramatically lower model scale when local degradation reasoning replaces brute-force capacity.

The baseline landscape further contextualizes this finding. FastDVDnet, the lightest baseline, achieves high throughput but sacrifices over 3\,dB in PSNR relative to UHD-GPGNet, confirming that speed-oriented design alone is insufficient for high-fidelity UHD denoising. Conversely, the heavier restorers RVRT and TAP$^\dagger$ do not translate their substantially larger parameter budgets ($12.8$M and $34.5$M, respectively) into better UVG quality under the matched protocol, suggesting that scaling temporal-modeling capacity without explicit degradation awareness yields diminishing returns.

\begin{figure*}[t]
\centering
\includegraphics[width=\textwidth]{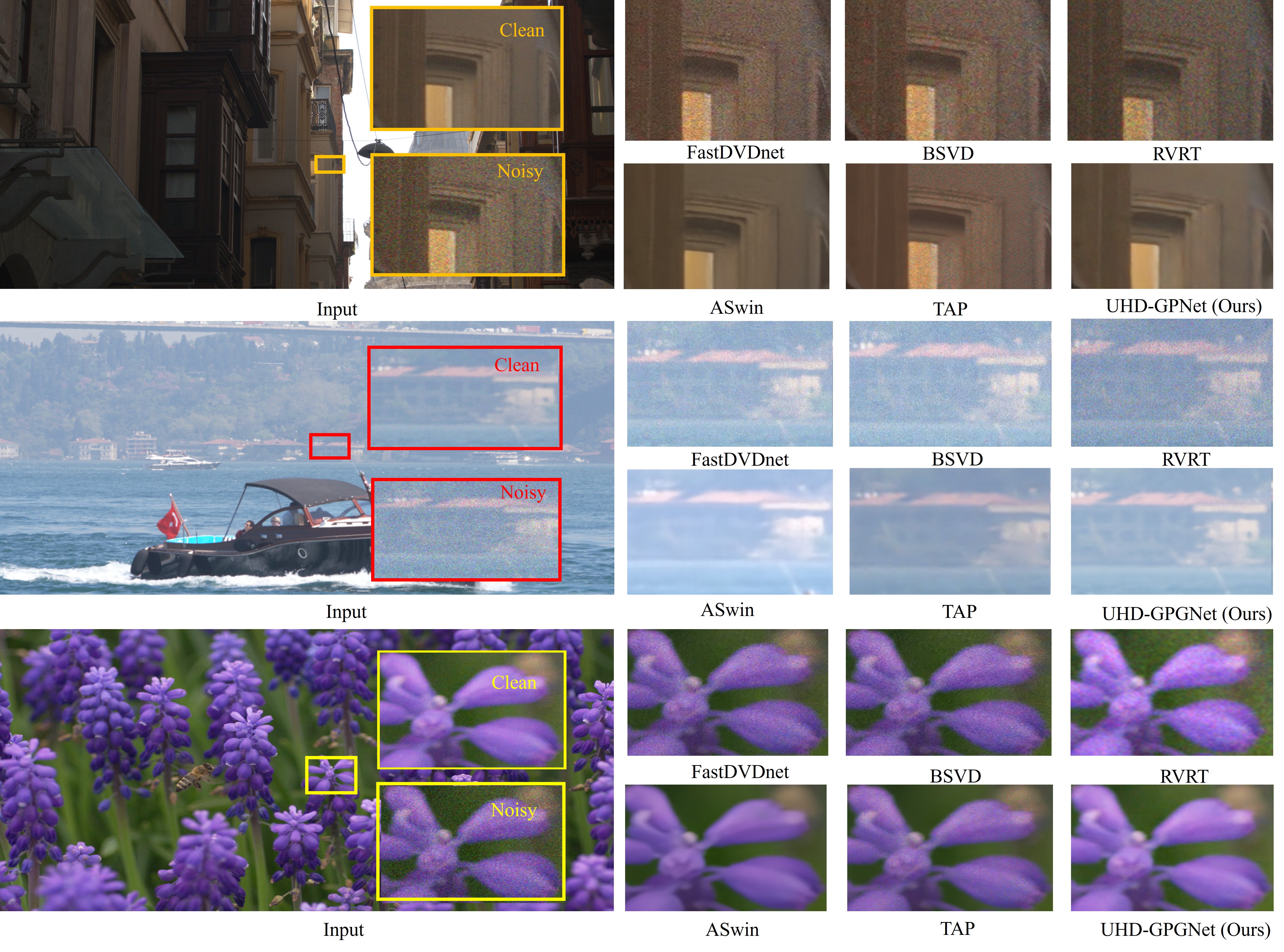}
\caption{Main qualitative comparison on UVG with one full 4K frame and enlarged crops emphasizing thin structures, repeated textures, and chroma-sensitive regions.}
\label{fig:main_qualitative}
\end{figure*}

Fig.~\ref{fig:main_qualitative} provides visual evidence consistent with these metrics. UHD-GPGNet suppresses residual grain more cleanly than the speed-oriented baselines in flat regions, avoids the over-smoothing visible on thin structures and repeated textures, and preserves sharper color transitions at chroma-sensitive boundaries. The deployment implications of this quality--complexity advantage are examined in Section~\ref{sec:deployment}.

\subsection{Controlled Cross-Scene Robustness on RealisVideo-4K}

To evaluate robustness beyond a single scene distribution, we construct a controlled five-point degradation schedule on RealisVideo-4K. Table~\ref{tab:degradation_protocol} defines three anchor settings ($\sigma=1.0$, $2.0$, $3.0$) that jointly scale exposure variation, sensor/grain noise, temporal flicker, chroma drift, and H.264 compression; intermediate points $\sigma=1.5$ and $2.5$ follow the same monotonic scaling. Because the protocol couples multiple physically motivated corruption types rather than varying a single factor, it tests robustness under practical mixed-degradation conditions.

\begin{table}[t]
\caption{Mixed-degradation protocol at three anchor settings; $\sigma{=}1.5$, $2.5$ follow the same monotonic scaling.}
\label{tab:degradation_protocol}
\centering
\scriptsize
\setlength{\tabcolsep}{2.5pt}
\renewcommand{\arraystretch}{1.05}
\begin{tabular}{@{}llll@{}}
\toprule
Component & $\sigma{=}1.0$ & $\sigma{=}2.0$ & $\sigma{=}3.0$ \\
\midrule
Exposure & 0.95--1.05 & 0.90--1.10 & 0.85--1.15 \\
\makecell[l]{Sensor+\\grain} & \makecell[l]{shot 1.5e-3--1.2e-2\\read 4e-4--4e-3\\$g_\sigma$ 8e-4--8e-3} & \makecell[l]{shot 3e-3--2.4e-2\\read 8e-4--8e-3\\$g_\sigma$ 1.6e-3--1.6e-2} & \makecell[l]{shot 4.5e-3--3.6e-2\\read 1.2e-3--1.2e-2\\$g_\sigma$ 2.4e-3--2.4e-2} \\
Flicker & \makecell[l]{$p{=}0.20$; $\rho{=}$0.85--0.97\\std${=}$(0.02, 0.003)} & \makecell[l]{$p{=}0.40$; $\rho{=}$0.85--0.97\\std${=}$(0.04, 0.006)} & \makecell[l]{$p{=}0.60$; $\rho{=}$0.85--0.97\\std${=}$(0.06, 0.009)} \\
Chroma & Cb/Cr std${=}$0.002 & Cb/Cr std${=}$0.004 & Cb/Cr std${=}$0.006 \\
H.264 & CRF${=}$28 & CRF${=}$32 & CRF${=}$36 \\
\bottomrule
\end{tabular}
\end{table}

\begin{table}[t]
\caption{Cross-scene robustness on RealisVideo-4K under five-point mixed degradation. PSNR at anchor settings $\sigma{=}1.0$, $2.0$, $3.0$; average SSIM over all five points; PSNR drop from $\sigma{=}1.0$ to $3.0$.}
\label{tab:realis_robustness}
\centering
\small
\setlength{\tabcolsep}{4pt}
\begin{tabular}{lccccc}
\toprule
Method & $\sigma=1.0$ & $\sigma=2.0$ & $\sigma=3.0$ & Avg.\ SSIM & Drop $1\!\rightarrow\!3$ \\
\midrule
FastDVDnet         & 33.43 & 29.17 & 26.58 & 0.768 & 6.85 \\
ASwin              & 36.23 & \textbf{32.37} & 29.50 & \textbf{0.962} & 6.73 \\
BSVD               & 32.29 & 28.53 & 26.18 & 0.718 & 6.11 \\
RVRT               & 32.43 & 28.52 & 27.14 & 0.734 & \textbf{5.29} \\
\midrule
TAP$^\dagger$      & 35.30 & 31.02 & 28.33 & 0.910 & 6.97 \\
\midrule
\textbf{UHD-GPGNet} & \textbf{36.43} & 32.14 & \textbf{30.22} & 0.951 & 6.21 \\
\bottomrule
\end{tabular}
\end{table}

\begin{figure}[t]
\centering
\includegraphics[width=\columnwidth]{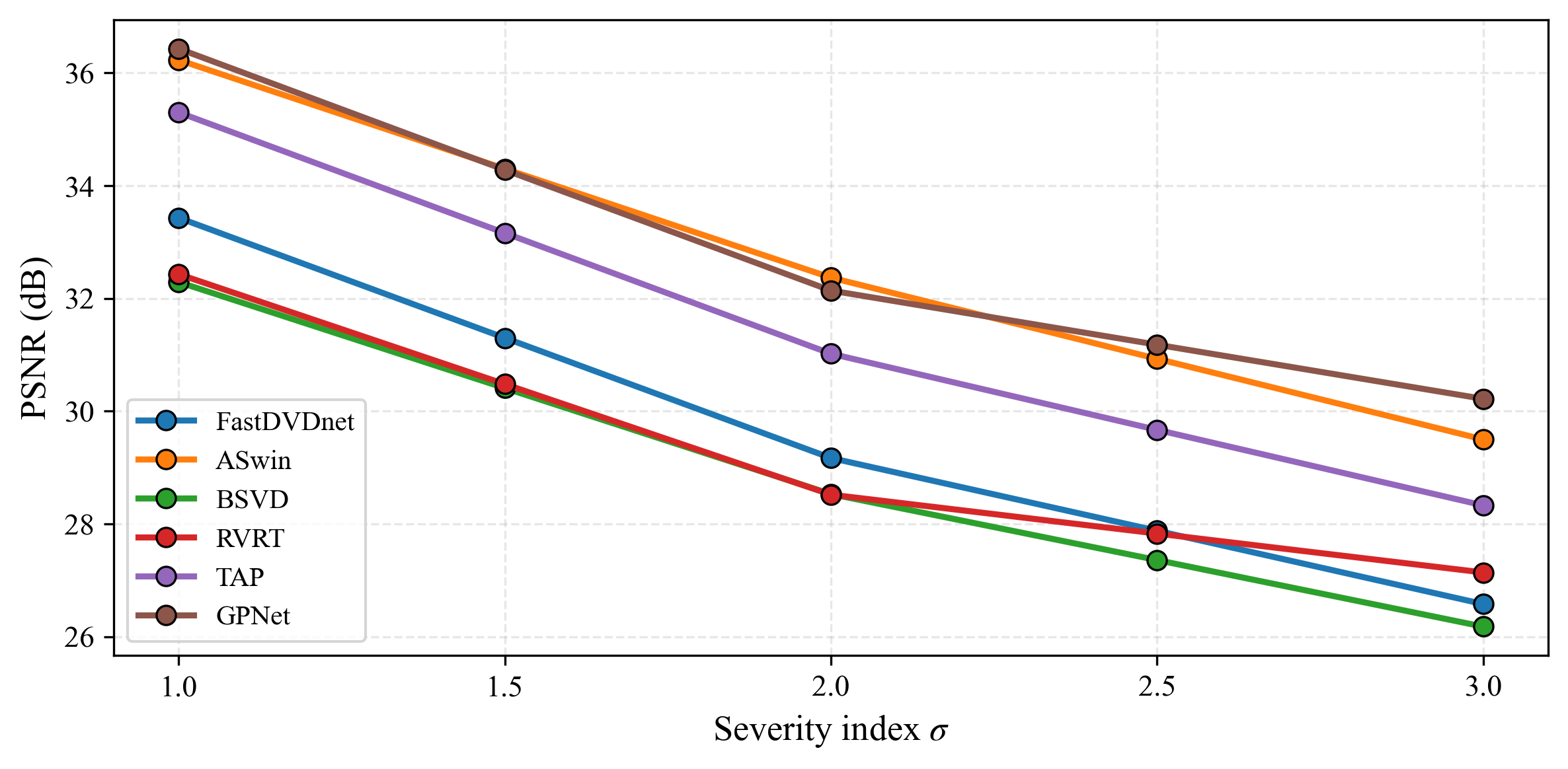}
\caption{RealisVideo-4K robustness curves under controlled mixed degradations at $\sigma=1.0$, $1.5$, $2.0$, $2.5$, and $3.0$.}
\label{fig:severity_curve}
\end{figure}

\begin{figure}[t]
\centering
\includegraphics[width=\columnwidth]{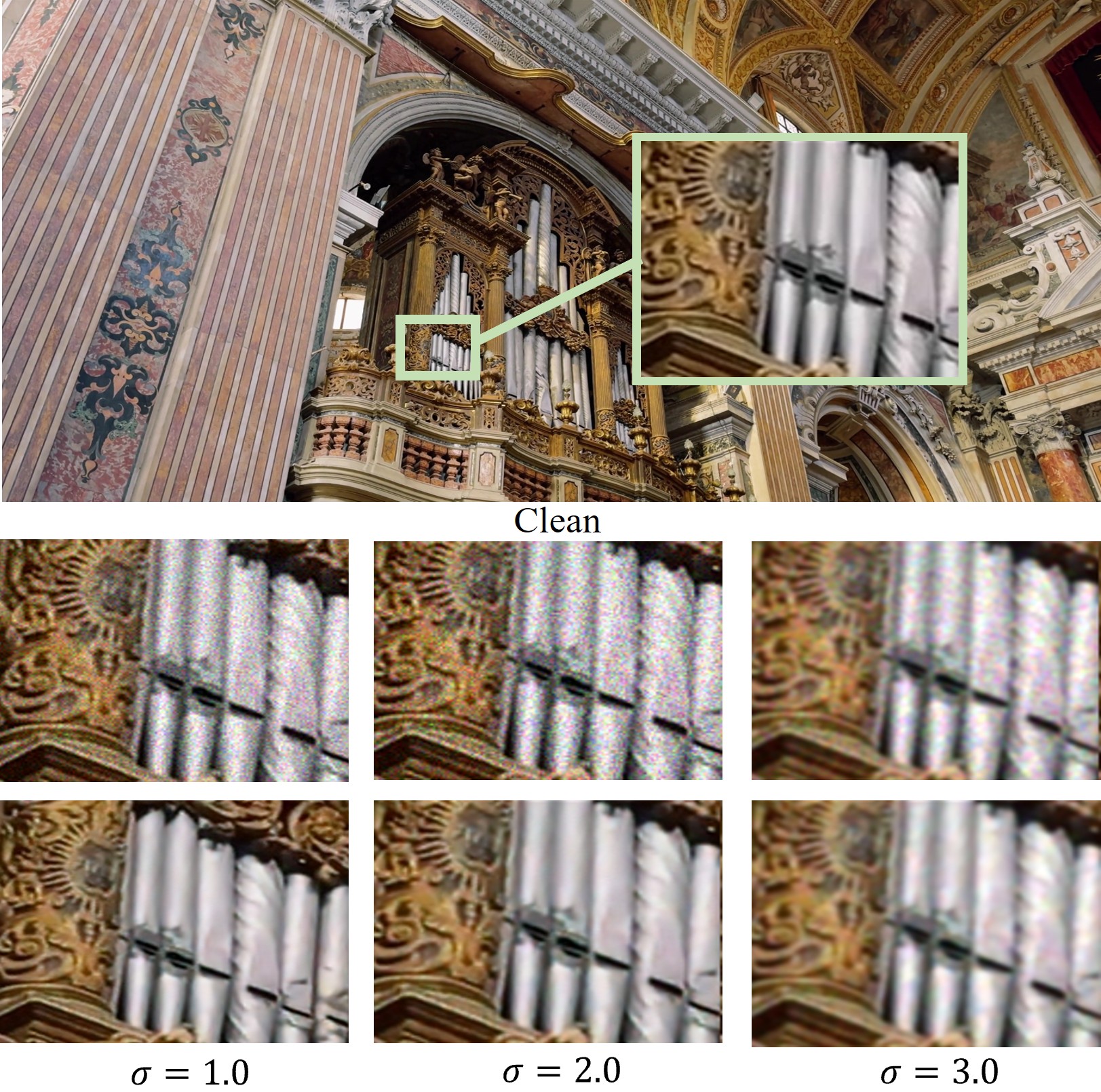}
\caption{Representative 4K crop visualization on RealisVideo-4K at $\sigma=1.0$, $2.0$, and $3.0$}
\label{fig:realis_crop}
\end{figure}

Table~\ref{tab:realis_robustness} and Fig.~\ref{fig:severity_curve} show that UHD-GPGNet achieves the highest PSNR at $\sigma{=}1.0$, $2.5$, and $3.0$, while ASwin holds a marginal edge at $\sigma{=}1.5$ and $2.0$. This crossover reflects different operating regimes: ASwin leverages window-attention capacity under mild corruption, whereas UHD-GPGNet becomes progressively more favorable as degradation intensifies and uncertainty cues provide greater benefit for fusion control. RVRT exhibits the smallest PSNR drop ($1{\to}3$: 5.29\,dB) but remains below both methods in absolute terms. ASwin retains a modest SSIM advantage (0.962 vs.\ 0.951), narrowing under heavier corruption (Fig.~\ref{fig:realis_crop}).

\subsection{Full-Resolution 4K Deployment Analysis}
\label{sec:deployment}
Table~\ref{tab:deployment_4k} profiles all methods under synchronized full-resolution $3840\times2160$ execution. ASwin processes the full input in one pass, accounting for its higher memory and latency.

The results separate the methods into three regimes. FastDVDnet achieves the highest throughput (62.52 FPS) but incurs a 4\,dB PSNR deficit. ASwin, RVRT, and TAP$^\dagger$ achieve stronger fidelity but at prohibitive latency: ASwin runs at only 0.95 FPS with 7.31 GiB peak memory. UHD-GPGNet combines the best of both regimes: it achieves the highest PSNR \emph{and} SSIM among all methods (34.213\,dB / 0.9639) while running at 28.06 FPS---$29.5\times$ faster than ASwin---with roughly half the peak memory. This operating point demonstrates that near-frontier restoration quality and practical 4K throughput are simultaneously achievable.At 28 FPS, UHD-GPGNet approaches the throughput needed for offline 4K post-production pipelines while maintaining only 3.95 GiB peak memory---less than a single consumer GPU's capacity---making it directly deployable without multi-GPU setups or model parallelism.

\begin{table}[t]
\caption{Full-resolution 4K deployment on RealisVideo-4K ($3840{\times}2160$) with synchronized measurement.}
\label{tab:deployment_4k}
\centering
\small
\setlength{\tabcolsep}{4pt}
\begin{tabular}{lccccc}
\toprule
Method & Peak Mem.\ (GiB)$\downarrow$& FPS$\uparrow$ & PSNR$\uparrow$ & SSIM$\uparrow$ \\
\midrule
FastDVDnet         & \textbf{1.32}    & \textbf{62.52} & 30.250 & 0.7714 \\
ASwin              & 7.31                   & 0.95           & 34.177 & 0.9631\\
BSVD               & 3.29                      & 12.95          & 29.640 & 0.7279 \\
RVRT               & 5.20                     & 2.04           & 29.485 & 0.7360 \\
\midrule
TAP$^\dagger$      & 4.94                     & 3.33           & 32.496 & 0.9107 \\
\midrule
\textbf{UHD-GPGNet} & 3.95                     & 28.06           & \textbf{34.213} & \textbf{0.9639} \\
\bottomrule
\end{tabular}
\end{table}

\subsection{Component Contribution and Mechanism Validation}

Table~\ref{tab:ablation_components} measures component contribution while Table~\ref{tab:mechanism_validation} validates mechanism behavior on specific failure modes.

The component progression shows that deterministic gating alone yields limited gain (31.95 vs.\ 31.78\,dB). The ``Attn+Var'' variant---which replaces the GP kernel with standard dot-product attention while retaining inducing-level variance prediction---improves to 32.11\,dB, confirming that explicit variance modeling is beneficial. A natural concern is whether the remaining gap to the full GP module (32.32\,dB) merely reflects the three extra kernel parameters ($\alpha$, $\ell$, $\gamma$). To control for this, we test ``Attn+Var+P,'' which augments dot-product attention with three matched learnable scalars (temperature, output scale, temporal weight) while keeping all other components identical. This parameter-matched control reaches only 32.15\,dB---a marginal +0.04\,dB over Attn+Var---while the GP-structured module achieves 32.32\,dB (+0.17\,dB beyond the control). The 0.17\,dB residual gap under matched parameter count confirms that the gain stems from the factored spatio-temporal RBF kernel and its induced locality structure, not from additional capacity. The remaining modules provide consistent refinements to 32.60\,dB. Table~\ref{tab:mechanism_validation} confirms that each design element addresses a distinct failure mode: removing GP causes the largest deterioration in error--uncertainty correlation ($0.75 \to 0.52$); removing the heteroscedastic objective weakens calibration; the RGB-only variant drops CbCr PSNR by 2.25\,dB; and disabling overlap raises seam score by $4.7\times$.

\begin{figure}[t]

\centering
\includegraphics[width=\columnwidth]{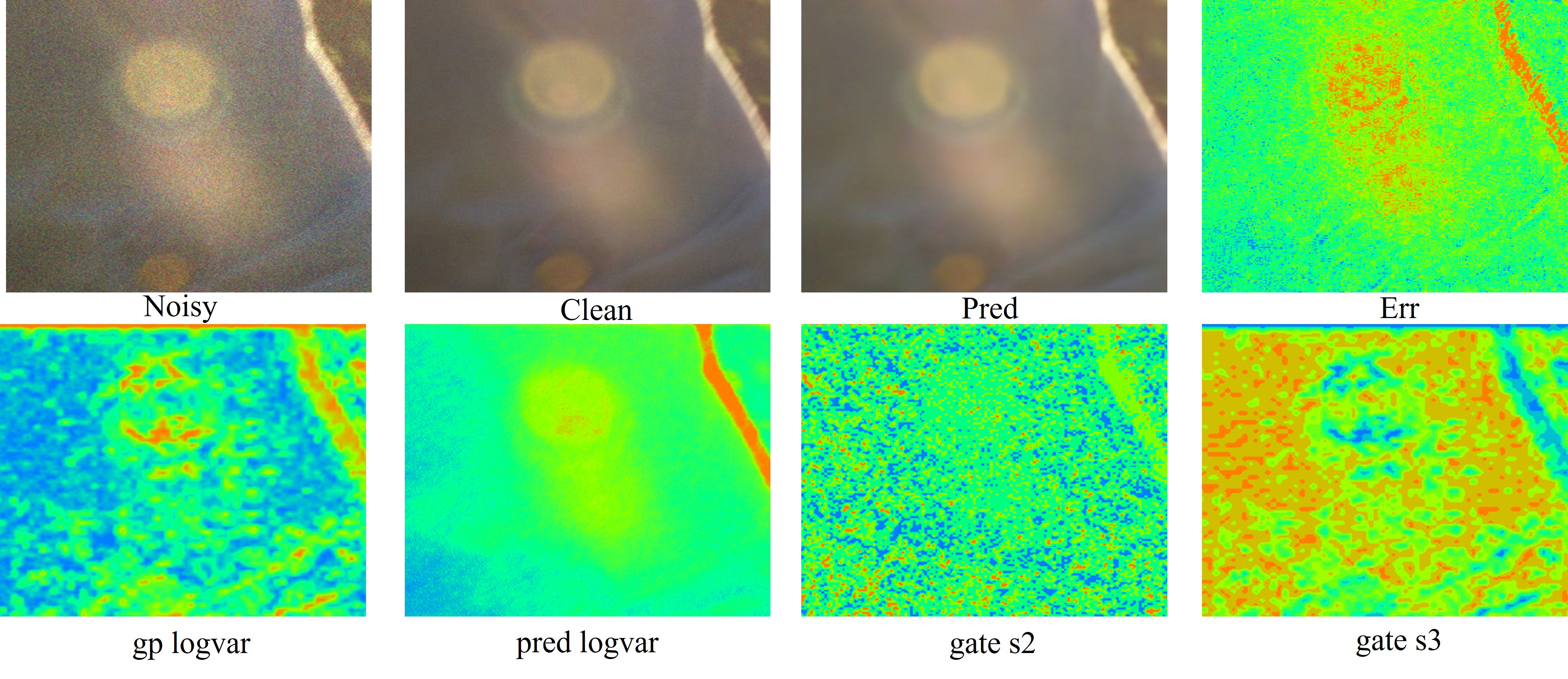}
\caption{Mechanism visualization: restored output, residual error, GP log-variance, predicted log-variance, and stage-2/3 fusion gates on a flare- and edge-contaminated region.}
\label{fig:mechanism_vis}

\end{figure}

Figure~\ref{fig:mechanism_vis} corroborates these findings: GP log-variance and predicted variance co-localize with residual error, and the stage-2/3 gates show a coarse-to-fine division of labor.

\begin{table}[t]
\caption{Component ablation. ``Attn+Var'' replaces the GP kernel with dot-product attention while retaining variance prediction. ``Attn+Var+P'' further adds three matched learnable scalars (temperature, output scale, temporal weight) to control for parameter-count differences with the GP kernel ($\alpha$, $\ell$, $\gamma$).}
\label{tab:ablation_components}
\centering
\small
\setlength{\tabcolsep}{4pt}
\begin{tabular}{lccccccc}
\toprule
Variant & Det.\ gate & GP & Hetero & Y/C & HF & PSNR$\uparrow$ & SSIM$\uparrow$ \\
\midrule
Backbone       & $\times$     & $\times$     & $\times$     & $\times$     & $\times$     & 31.78 & 0.881 \\
+ Det.\ gate   & $\checkmark$ & $\times$     & $\times$     & $\times$     & $\times$     & 31.95 & 0.885 \\
+ Attn+Var     & $\times$     & attn         & $\times$     & $\times$     & $\times$     & 32.11 & 0.891 \\
+ Attn+Var+P   & $\times$     & attn+p       & $\times$     & $\times$     & $\times$     & 32.15 & 0.893 \\
+ Sparse GP    & $\times$     & $\checkmark$ & $\times$     & $\times$     & $\times$     & 32.32 & 0.898 \\
+ Hetero obj.  & $\times$     & $\checkmark$ & $\checkmark$ & $\times$     & $\times$     & 32.41 & 0.902 \\
+ Y/C head     & $\times$     & $\checkmark$ & $\checkmark$ & $\checkmark$ & $\times$     & 32.49 & 0.905 \\
Full (+HF)     & $\times$     & $\checkmark$ & $\checkmark$ & $\checkmark$ & $\checkmark$ & 32.60 & 0.910 \\
\bottomrule
\end{tabular}
\end{table}

\begin{table}[t]
\caption{Mechanism validation of uncertainty quality, chroma stability, chroma fidelity, and tiled deployment behavior.}
\label{tab:mechanism_validation}
\centering
\footnotesize
\setlength{\tabcolsep}{3.5pt}
\begin{tabular}{@{}lccccc@{}}
\toprule
Setting & \makecell{Err--Unc\\Corr.$\uparrow$} & \makecell{Chroma\\Flicker$\downarrow$} & \makecell{CbCr\\PSNR$\uparrow$} & \makecell{Seam\\Score$\downarrow$} & \makecell{Ref.\\PSNR$\uparrow$} \\
\midrule
No GP       & 0.52 & 0.0234 & 37.19 & 0.0121 & 32.18 \\
No hetero   & 0.59 & 0.0198 & 37.85 & 0.0113 & 32.29 \\
RGB-only    & 0.61 & 0.0231 & 36.87 & 0.0107 & 32.21 \\
No overlap  & 0.61 & 0.0185 & 37.96 & 0.0292 & 32.44 \\
Full model  & \textbf{0.75} & \textbf{0.0125} & \textbf{39.12} & \textbf{0.0062} & \textbf{32.60} \\
\bottomrule
\end{tabular}
\end{table}

\subsection{Real-World Generalization and Downstream Detection}
\label{sec:realworld}

The preceding experiments use matched synthetic degradation. To test generalization to unseen real sensor noise, we apply the frozen pretrained UHD-GPGNet to 4K video captured with different smartphone(iPhone 16 Pro Max, Huawei Mate 70 and Samsung Galaxy S25) under challenging low-light and indoor conditions, without retraining.

Because no clean ground truth is available, we evaluate through (i)~\emph{downstream detection improvement} using an off-the-shelf YOLOv8 detector (detection count, confidence, cross-frame consistency) and (ii)~\emph{no-reference perceptual quality} via NIQE and BRISQUE (lower is better).

Table~\ref{tab:realworld} reports averaged results across scenes spanning outdoor dusk streets and indoor corridors. Denoising consistently improves all indicators: detection count by $+0.3$/frame, confidence by $+0.009$, temporal consistency from $0.863$ to $0.878$, NIQE from $3.66$ to $3.55$, and BRISQUE from $19.8$ to $19.0$.

\begin{figure}[t]
\centering
\includegraphics[width=\columnwidth]{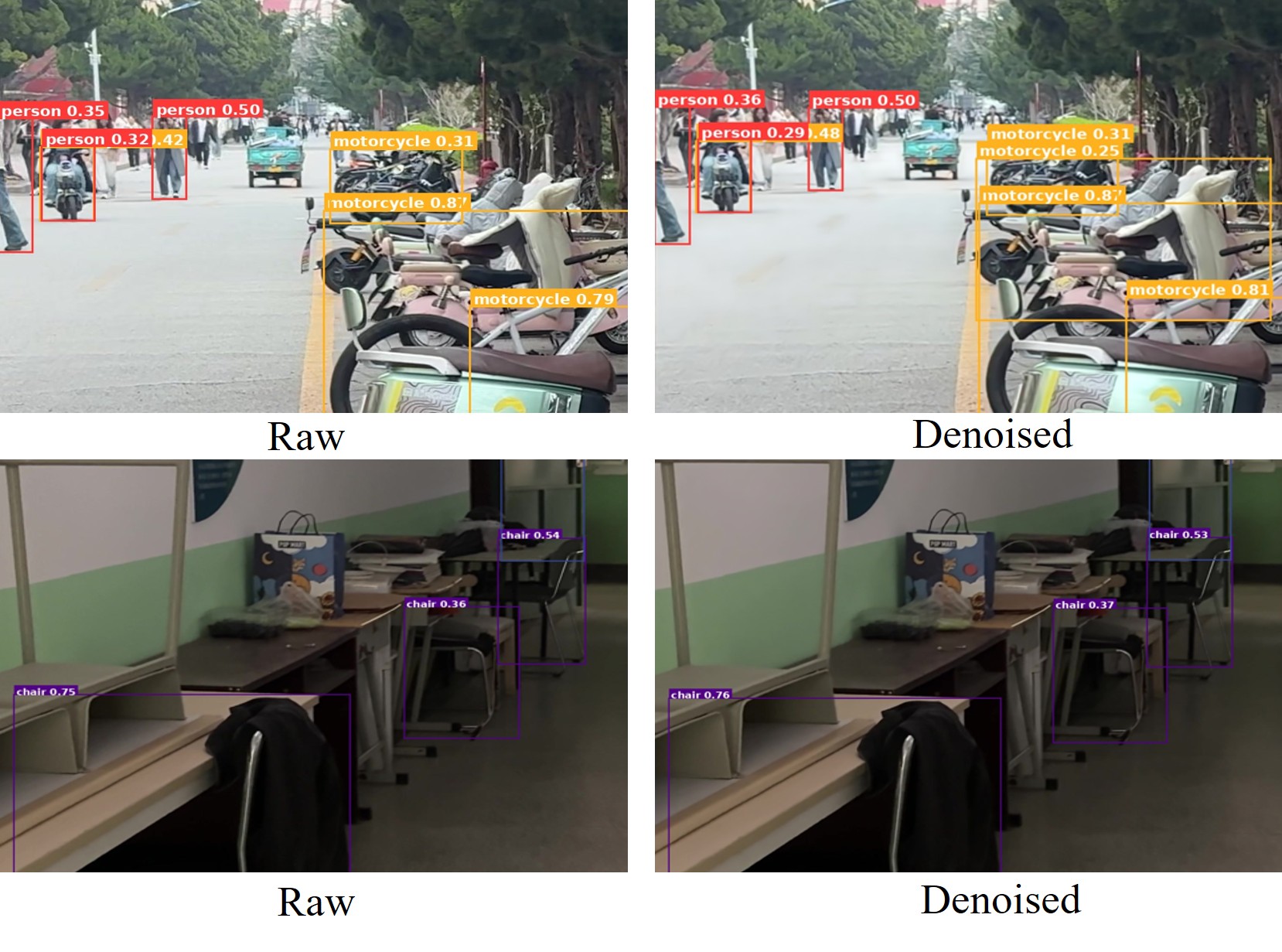}
\vspace{-3mm}
\caption{Real-world downstream detection on phone-captured 4K video. \textbf{Top}: outdoor dusk scene---denoising recovers distant targets and increases detection confidence. \textbf{Bottom}: indoor low-light scene---object boundaries become sharper and detection boxes stabilize. The model is applied without retraining on real noise.}
\label{fig:realworld_det}
\vspace{-4mm}
\end{figure}

Fig.~\ref{fig:realworld_det} shows representative evidence: the denoised outdoor frame recovers distant targets missed on the noisy input, and the indoor frame exhibits sharper boundaries with higher detection confidence. These results confirm that fidelity gains from synthetic training transfer to real sensor noise and yield measurable downstream perception improvements.

\begin{table}[t]
\caption{Real-world phone-captured 4K evaluation (averaged across scenes). The frozen pretrained model is applied without retraining. Detection uses YOLOv8; quality uses no-reference metrics (lower = better).}
\label{tab:realworld}
\centering
\footnotesize
\setlength{\tabcolsep}{4pt}
\begin{tabular}{@{}lcccccc@{}}
\toprule
& \makecell{Avg\\Det$\uparrow$} & \makecell{Avg\\Conf$\uparrow$} & \makecell{Temp\\Cons$\uparrow$} & \makecell{NIQE\\$\downarrow$} & \makecell{BRISQUE\\$\downarrow$} \\
\midrule
Raw input       & 6.2 & 0.414 & 0.863 & 3.66 & 19.8 \\
\textbf{Denoised} & \textbf{6.5} & \textbf{0.423} & \textbf{0.878} & \textbf{3.55} & \textbf{19.0} \\
\midrule
$\Delta$ & \textcolor{black}{+0.3} & \textcolor{black}{+0.009} & \textcolor{black}{+0.015} & \textcolor{black}{$-$0.11} & \textcolor{black}{$-$0.8} \\
\bottomrule
\end{tabular}

\vspace{-4mm}
\end{table}

\section{Discussion and Limitations}

\noindent \textbf{GP-guided vs.\ attention-based fusion.}
UHD-GPGNet matches ASwin's PSNR with $12.6\times$ fewer parameters on UVG and surpasses it in both PSNR and SSIM under full-resolution 4K deployment at $29.5\times$ higher throughput. On RealisVideo-4K, it surpasses ASwin at heavier degradation levels while ASwin retains a marginal edge under milder corruption. This crossover suggests complementary strengths: attention aggregates efficiently when structure is accessible, whereas GP-guided uncertainty cues become critical when local corruption heterogeneity is high. Combining both mechanisms---e.g., GP-guided gating within a lightweight attention backbone---is a promising direction.

\noindent \textbf{Evaluation scope and temporal consistency.}
The controlled experiments (Sections~4.2--4.5) use a matched synthetic degradation protocol, while Section~\ref{sec:realworld} complements them with a real-world generalization study on phone-captured 4K video under unmatched noise conditions. Together, these two evaluation axes cover both controlled reproducibility and practical applicability. The mixed-degradation protocol does not isolate individual degradation types; disentangling sensitivities to sensor noise, compression, and chroma drift individually is left to future work. Regarding temporal quality, the current evaluation captures temporal behavior through chroma flicker reduction (Table~\ref{tab:mechanism_validation}: $0.0125$ vs.\ $0.0234$ without GP) and cross-frame detection consistency in the real-world study (Table~\ref{tab:realworld}: $0.863 \to 0.878$).
% ; dense per-frame temporal metrics (tOF, tLP~\cite{teed2020raft}) and perceptual measures (LPIPS~\cite{zhang2018perceptual}) beyond PSNR/SSIM~\cite{wang2004image} would further enrich the assessment.

\noindent \textbf{Modeling and deployment scope.}
The sparse GP module adds modest overhead ($M{=}16$ inducing tokens) but operates locally within each stage on five-frame clips, limiting the temporal uncertainty horizon. Scenes with longer-range dependencies may benefit from recurrent or hierarchical uncertainty propagation. The overlap-tiled inference trades seam quality against throughput; hardware-adaptive strategies that adjust tile size based on available memory would broaden practical applicability beyond the single-GPU setting studied here.

\section{Conclusion}

We presented UHD-GPGNet, a practical UHD video denoising framework where sparse GP-guided local posterior estimation provides explicit degradation cues for adaptive temporal--detail fusion without dense probabilistic inference. Experiments demonstrate the best restoration quality among all compared methods at $12.6\times$ lower parameter cost and $29.5\times$ higher 4K throughput than the strongest fidelity baseline, robust performance across a five-point mixed-degradation schedule, and a favorable quality--speed operating point for full-resolution 4K deployment. A real-world study on phone-captured 4K video further shows that the model generalizes to unseen sensor noise and improves downstream object detection, confirming the practical applicability claimed by the design. Ablation confirms that GP-guided modeling---not generic capacity---drives the principal gain, while auxiliary designs address specific UHD failure modes. The results suggest that sparse local probabilistic reasoning offers a principled, deployable alternative to purely implicit fusion for UHD video restoration.

\section*{Acknowledgments}
The authors thank the anonymous reviewers for their constructive feedback. This work was supported in part by [funding source]. The authors declare no competing interests.

%%
%% The next two lines define the bibliography style to be used, and
%% the bibliography file.
\bibliographystyle{ACM-Reference-Format}
\bibliography{GPUnet_ref}

% Appendix removed -- no supplementary content.

\end{document}